# Evaluation of Coreference Rules on Complex Narrative Texts

*Andrei Popescu-Belis and Isabelle Robba*

**Abstract**

This article studies the problem of assessing relevance to each of the rules of a reference resolution system. The reference solver described here stems from a formal model of reference and is integrated in a reference processing workbench. Evaluation of the reference resolution is essential, as it enables differential evaluation of individual rules. Numerical values of these measures are given, and discussed, for simple selection rules and other processing rules; such measures are then studied for numerical parameters.

## 1. Introduction

Reference resolution advancements in the past years have shown that it has become a task which can be performed with interesting performance levels even on unrestricted texts. The Message Understanding Conferences (hence, MUC: ARPA 1995) have proposed a coreference resolution task related to a "markable recognition" task: the participants had to mark referring expressions (RE) in the given texts as well as coreference links between them. As the texts belonged to restricted domains – further tasks being strongly domain-dependent – the systems used domain-specific knowledge to solve coreference, as for instance LaSIE (Gaizauskas & al. 1995), together with more general cues.

In any case, it seems particularly interesting for developers and evaluators to be able to assess quantitatively the role of each knowledge source in coreference resolution. In other words, if the contribution of each piece of knowledge to the solving process can be measured, then one could evaluate the performance of more sophisticated heuristics compared to the basic ones, and decide, for instance, which ones are worth implementing in a given system. This supposes of course that experimental results on a system are available to the whole research community; we outline in this paper a possible description for such data, giving our first experimental assessment of some simple coreference rules, in the model-based reference processing system that we implemented.

A brief overview of our system and connected tools will be given first (§2.1), followed by a short description of the theoretical model of reference (§2.2). We then describe our reference solver, the algorithm and knowledge it uses (§2.3). The narrative texts used for evaluation have their own importance (§2.4) and even more the original scoring methods we employ (§2.5), which are compared to more classical ones and other possibilities. At this stage, a natural way to assess relevance of coreference rules is presented (§3.1) and applied successfully to rank our set of selection rules (§3.2). The same sort of assessment discards rules for definite vs. indefinite NPs (§3.3). Finally, we discuss extensions of these considerations to numeric parameters of our solver (§3.4).

## 2. Overview of the reference resolution workbench

### 2.1 A modular architecture

One cannot approach reference resolution on a higher scale unless several processing tools and resources are available. First of all, texts have to be brought to a specific format, either manually or automatically; also, the referring expressions (REs) designating discourse referents have to be isolated from the text, by specific mark-up for instance. Evaluation of the solver is a crucial step (the notion of 'improvement' itself depends on it), and in order to perform it a correct solution has to be defined for the given text. This solution (key) is sent to an evaluator, together with the system's response, to obtain a score for the system.

We implemented these modules (Figure 1) on a workstation, and used SGML tags to mark up REs as well as reference information according to our model of the phenomenon. Our converters accept various SGML-based input and output formats, as described in (Popescu-Belis 1998). An external parser attaches an f-structure (parse tree in LFG) to each RE. The 'Processing Units' are passed on to the reference solver, which produces the response discourse referents (or MRs, see next §), and further on to the evaluator. An optimizer can control this chain to tune the solver's parameters for a given text (Popescu-Belis 1998).

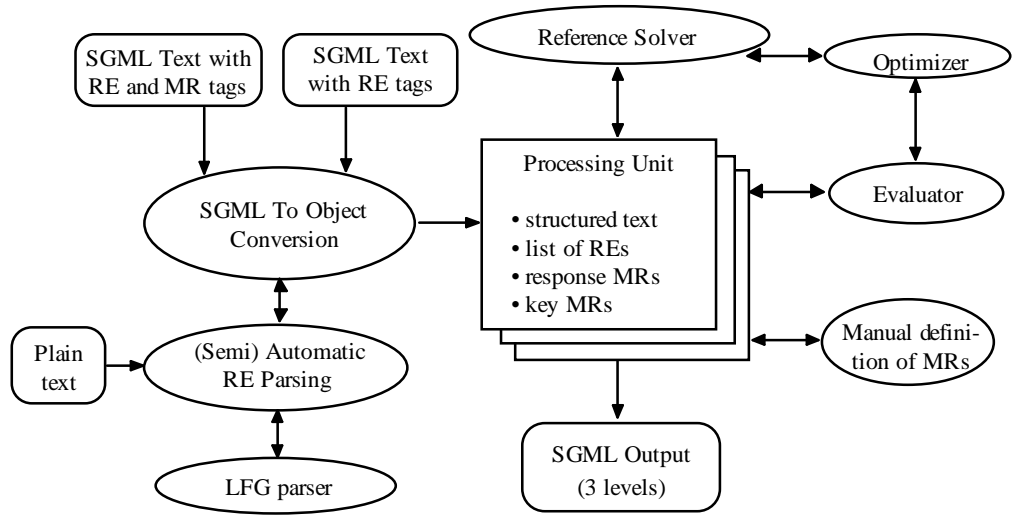

Figure 1. General architecture of the reference resolution workbench

## 2.2 The mental representation (MR) paradigm

The theoretic model on which the reference solver is based has been described elsewhere (Popescu-Belis, Robba & Sabah 1998). To sum up, the genuine goal of reference resolution is believed to be the construction of *"mental" representations (MRs)* of the discourse referents, and not only the identification of links between them. In the case of narrative texts understanding, the MRs are to be build using the text itself, and particularly its referring expressions (REs). World knowledge is useful, if available, but unlike other possible tasks, there is scarcely any perceptual information to be used here.

These theoretical considerations were of course adapted to the available technical capacities. We thus designed a relatively low-knowledge system (cf. next §) while still remaining within the MR paradigm. Our MR data structures are sets of REs, and though there is still no semantic structure for the MR itself, its features are obtained by examining and averaging the features of the composing REs. Also, the system maintains an activation value (number) for each MR, which models the MR's salience to the mind of a human reader of the narrative text; again, this feature is specific to the MR (or the set of REs).

These remarks make it quite clear that the MR paradigm is not reducible to simple coreference, i.e., pairing of REs; indeed, even if tracking coreference links enables one to build MRs, this construction is only *a posteriori*, and misses the dynamic properties of the MR paradigm, activation and 'feature averaging'. Conversely, these properties are also present even in our implementation of the weaker MR paradigm (low knowledge). Still more different is the antecedent/anaphor paradigm, which builds asymmetrical coreference links, and separates REs in two artificial classes. Following Ariel (1990) and Reboul (1994) there seems to be rather a continuum between the two classes, more and more knowledge from the text being needed to solve the RE (maximal for pronouns). But, basically, the resolution task is the same, i.e. processing the current RE, as shown below.

## 2.3 The reference resolution algorithm

The reference resolution mechanism has been implemented using an algorithm in the style of Lappin and Leass (1994) which models the selection/activation paradigm. The REs are examined sequentially, and for each of them, the solver either associates it to an existing MR (if the RE refers to an already introduced discourse referent), or it creates a new MR for it. To do this, the solver proceeds in two steps. First, it determines the set of MRs to which the RE could possibly be attached. Then, it selects in this set the MR possessing the highest activation value, and associates the current RE to it. If the set is empty, a new MR is created. Afterwards, activation is updated for all MRs.

For the first step, three selectional constraints are examined between the current RE and each RE. Two of them are morpho-syntactic, they check number and gender agreement between REs. The third is semantic and uses a semantic network. It examines compatibility between heads and modifiers of the NPs in the two REs. Actually, this network is limited to the concepts present in the studied corpus, but it would be replaced with benefit by a WordNet type network for French.

Several methods have been implemented to combine selectional constraints for an MR; that is, if an MR consists of $\{RE_1, RE_2, RE_3 \ldots RE_n\}$, the system has to find out whether the current RE can be attached to the MR on the basis of compatibility between RE and all/some of the $RE_i$. Here are some heuristics:

- H1: the constraints are applied to $RE_1$ (but the first RE referring to an object isn't always the most typical);
- H2: the constraints are applied to all non-pronominal REs;
- H3: the constraints are applied to each RE until the solver finds a non-pronominal RE which respects these constraints;
- H4: the constraints are applied to each RE, and if X% of them satisfy them, then the MR does too.

The activation value accompanying each MR reflects the salience of the referent all along the text. When a new MR is created, or when an RE is attached to an existing MR, the activation of the MR is recalculated according to the nature of the RE. Also, the activation of an MR decreases in time, and there is a limited size buffer storing only the most active MRs, the other being archived. More details about the parameters are given at §3.4.

## 2.4 The texts

Two French texts have been used to test the reference solver. The first one is a short story by Stendhal, *Vittoria Accoramboni* (VA), and the second is the first chapter of a novel by Balzac, *Le Père Goriot* (LPG). These texts were pre-processed before being tested. At LIMSI, VA underwent manual tagging of paragraphs, sentences and boundaries of all REs, then automatic conversion to Smalltalk objects (our programming language). Using Vapillon's & al. (1997) LFG parser, an f-structure (parse tree) was added to each RE. Then the correct MRs were created using our user-friendly interface.

LPG was already SGML-encoded at LORIA (Nancy) with RE and MR tags using Bruneseaux and Romary (1997) mark-up conventions. However, only the REs corresponding to the main characters of the first chapter were encoded: humans, places and objects. This explains why the ratio RE/MR is much greater for LPG than for VA (7.00 vs. 1.72).

The LPG text was also converted to Smalltalk objects, f-structures were added to the REs and MRs were automatically generated from the SGML tags. Table 1 shows the main characteristics of these two texts.

|        | VA   | LPG   |
|--------|------|-------|
| Words  | 2630 | 28576 |
| REs    | 638  | 3359  |
| MRs (key) | 372 | 480 |
| RE / MR | 1.72 | 7.00 |

| Nominal REs    | 510 | 1864 |
|----------------|-----|------|
| Pronoun REs    | 102 | 1398 |
| Not parsed REs | 26  | 97   |

Table 1. Characteristics of the two texts

## 2.5 Evaluation of reference resolution

When working on texts of a certain length, automatic scoring is necessary to evaluate reference resolution. Indeed, manual scoring is not only long and tedious as soon as texts become complex, it is *per se* a non-trivial theoretical problem. Even in the 'coreference view', counting correct and missing links in the system's response (namely, precision and recall) is clearly not enough: there are different ways of establishing links which correspond to the same 'understanding' of the text's referents. It has been, to our knowledge, the merit of Vilain & al. (1995) to design a scoring method that focuses not on coreference links between REs but on the groups of REs which represent the same discourse referent, the method was used for MUC-6 and 7 (ARPA 1995). In this view, evaluating reference resolution means comparing two partitions of the global RE set, the correct one (key) and the system's response. Vilain & al.'s algorithm compares these partitions indulgently, counting the minimal number of missing, resp. wrong links imputable to the system. As we have shown elsewhere (Popescu-Belis & Robba 1998) this can produce misleading results, especially for texts with high coreference rates as in our case: a trivial but efficient tactique is then to over-group REs into MRs. The MUC algorithm indulgency is also visible on Tables 2 and 3, where the bottom line corresponds to a single (huge) response MR.

We have proposed three new algorithms which attempt to overcome this sort of limitation (Popescu-Belis & Robba 1998). We will use here two of them, which rely also on 'recall' and 'precision', but count these scores after attributing to each key referent (or MR) a most probable corresponding response MR, called 'core MR'. Depending on whether two key MRs may or may not share the same core-MR, we obtain two scoring methods, 'core-MR' and 'ex-core-MR'. We have proven that 'core-MR' scores are always lower than those of the MUC-6 and have given theoretical and experimental evidence that our algorithms are more robust and less indulgent, and that is why we will use them in the next section.

This kind of evaluation isn't of course the only possible one for coreference resolution. There are matters which aren't easily handled by such algorithms, as for instance limited evaluation of only one class of REs, which is a more general view of anaphora resolution evaluation (namely, for the pronoun class). It may even seem, from this point of view, that anaphora resolution cannot be evaluated satisfyingly unless all key MRs are marked, because there is no reason to prefer antecedent A to antecedent B if they are actually coreferent.

Researchers in anaphora resolution have sometimes proposed other measures. R. Mitkov ((Mitkov & Belguith 1998), also personal communication) suggested that comparison of a system against a 'base-line model' using only very elementary knowledge could be more relevant than absolute recall and precision values. Indeed, it is essential to appreciate the exact contribution of specific (and possibly sophisticated) knowledge to the system's result, and the methods we present here (§3) try also to answer this question. It is however worth noting that evaluation against a base-line model still needs an evaluation algorithm (applied twice) but produces a relative, not absolute score.

## 3. Assessing the relevance of individual coreference rules

### 3.1 Motivation and measure definition

As stated in the introduction there are at least two tasks which benefit from a precise evaluation of the role of each coreference rule. First, this evaluation increases the understanding of a complex phenomena, in which thousands of REs are involved. In such a case, evaluation (§2.5) through precision and recall is far too short-sighted to provide understanding of the interaction among rules. If the role of individual rules can be assessed separately, then the developers can test whether the measured contribution of each rule corresponds to the one estimated on theoretical or empirical (non-experimental) grounds. If it does not, the system may need fixing, or the rule may have to be re-analyzed. Second, for systems which are developed from scratch, assessments obtained for other systems can orient work towards the most convenient rules, those that provide the highest increase in score with the most simple knowledge available.

The difficulty of such an evaluation lies of course in the fact that if a system needs, at a certain point, N rules, and obtains an S% score (be it recall or precision or other) one cannot count the contribution of each rule as a fraction of S, with the sum giving S. More formally, if the rules are $R_1$ to $R_N$, and $C(R_i)$ are their contributions to S (their relevance for the current text), there is of course no simple way to compute some $C(R_i)$ so that

$$\sum_{1\ldots N} C(R_i) = S$$

For instance, if $C_a(R_i)$ is the system's score with $R_i$ alone, and $C_m(R_i)$ the score when only $R_i$ is missing, it is clearly untrue that

$$\sum_{1\ldots N} C_a(R_i) = S \quad \text{or} \quad \sum_{1\ldots N} (S - C_m(R_i)) = S$$

That is because in general the rules are not independent, they mutually influence one another. For the set of selection rules of our system, these sums are given in the next §. Also, turning off a rule may increase one score (e.g. recall) while decreasing the global score (f-measure).

Our proposal is a first attempt to solve this difficulty, drawing precisely on both of the 'contributions' suggested above. We suppose that the score is one of the f-measures. For each rule, we compute the score decrease when the rule is removed, $S-C_m(R_i)$, and the score with the rule alone $C_a(R_i)$; then, instead of further computation, we only order the rules according to *both* of the coefficients . An empirical analysis suggests that these coefficients are indeed co-variant: if removing a rule induces an important decrease, then this rule alone induces a good score. Some figures are given in the next §.

This method is of course adapted to sets of quasi-independent rules; if for instance triggering R2 depends on triggering R1, then the coefficients are no more independent, but one could still consider R1-R2 as complex rule, and use the method.

### 3.2 Relevance of the selection rules

In a series of trials, we kept all the parameters constant and discarded alternatively the selection rules. Results are given Table 2 for VA and Table 3 for LPG: the f-measure is shown for the MUC 6-7 scoring method, as well as for two of our methods which compute recall and precision as well. As stated above, however, these two methods are less indulgent than the MUC one, and they show higher amplitude variations; that is why we give recall and precision for the 'Core MR' scoring method (Popescu-Belis & Robba 1998), the most relevant one. The first line shows the percentage value of the scores when the three selection rules are loaded; on the next lines, variations are given when one on three rules are discarded.

Assessment of the relative importance of the three selection rules $R_G$, $R_N$ and $R_S$ using these results is particularly coherent. It appears that removing any of the rules decreases the precision, which proves the relevance of each rule. As these rules prevent attachment of REs to incompatible MRs, removing them leads normally to a better recall (correct links being found among the numerous and random wrong ones): it is effectively the case for MUC-recall (not represented) and for Core-MR-recall on the LPG text — however, due to MR forgetting effects, this is not the case for the VA text, where recall sometimes decreases when rules are removed.

It is nevertheless clear that the most important increases in recall appear when the semantic compatibility filter $R_S$ is removed; this should of course be a rather positive result, weren't it symptomatic of MR over-grouping which ruins both precision and f-measure. Again, precision decrease is the most dramatic when $R_S$ is removed, whatever rules are left ($R_G$, $R_N$, or both) — compare precision and f-measures on lines 2, 3 an 5 with those on lines 4, 6 and 7 (the latter are more than three times the former). These results coherently rank $R_S$ as the most effective selection rule.

Among the other two selection rules, it is again clear (though less dramatic) that the gender agreement is more effective than the number agreement. Indeed, removing $R_G$ always induces greater score decrease than removing $R_N$, whether $R_S$ is present (line 2 vs. line 3) or absent (line 7 vs. line 6), again looking at precision and f-measure. Conversely, recall increase due to over-grouping is more important when $R_G$ is removed. Greater effectiveness of $R_G$ is easy to understand: the amount of plural reference (plural coreferent REs) is relatively small compared to the amount of REs for which gender is relevant, as in French even inanimate referents have gender.

| $R_G$ | $R_N$ | $R_S$ | MUC f-measure | Core-MR recall | Core-MR precision | Core-MR f-measure | ExCore-MR f-measure |
|---|---|---|---|---|---|---|---|
| x | x | x | **65.48** | **53.38** | **49.96** | **51.61** | **70.49** |
|   | x | x | − 0.43 | − 7.14 | − 9.04 | − 8.19 | − 4.98 |
| x |   | x | − 1.07 | − 5.64 | − 8.40 | − 7.17 | − 6.58 |
| x | x |   | − 7.88 | + 36.47 | − 32.55 | − 22.36 | − 49.41 |
|   |   | x | − 0.43 | − 5.26 | − 11.18 | − 8.66 | − 6.36 |
| x |   |   | − 7.27 | + 36.68 | − 33.57 | − 23.88 | − 50.43 |
|   | x |   | − 6.35 | + 43.99 | − 38.98 | − 31.86 | − 55.68 |
|   |   |   | − 6.56 | + 46.62 | − 40.23 | − 33.88 | − 60.62 |

Table 2. Results on the VA text — The first line shows scores (%) when the three selection rules are loaded; $R_G$ stands for gender agreement, $R_N$ for number and $R_S$ for semantic compatibility. Rules are then discarded (unticked) one by one, and score variations with respect to the first line are given for each combination.

| $R_G$ | $R_N$ | $R_S$ | MUC f-measure | Core-MR recall | Core-MR precision | Core-MR f-measure | ExCore-MR f-measure |
|---|---|---|---|---|---|---|---|
| x | x | x | **78.40** | **41.41** | **44.48** | **42.89** | **41.46** |
|   | x | x | + 0.26 | + 4.07 | − 1.91 | + 1.09 | − 0.40 |
| x |   | x | + 0.80 | + 0.25 | − 1.75 | − 0.70 | − 0.86 |
| x | x |   | + 8.45 | + 40.58 | − 19.63 | − 4.75 | − 14.83 |
|   |   | x | + 1.58 | + 4.94 | − 3.23 | + 0.76 | − 1.16 |
| x |   |   | + 9.20 | + 42.07 | − 20.82 | − 6.02 | − 15.99 |
|   | x |   | + 9.44 | + 49.17 | − 24.85 | − 10.62 | − 19.93 |
|   |   |   | + 13.87 | + 58.59 | − 26.79 | − 12.83 | − 23.79 |

Table 3. Results for the LPG text — Same legend as Table 2

## 3.3 Relevance of the rules for definite/indefinite NPs

The same data analysis can be applied to other rules of the reference solver, provided they have also binary values: for instance, a rule that can be active or discarded, or a flag which is or not set. We tested here simple rules which are grounded in a linguistic heuristic: it is generally observed that discourse referents are introduced in narrative texts by indefinite noun phrases, and often referred to using alternatively pronouns or definite NPs. This rule is by no means absolute, as some definite NPs also introduce MRs, but as an heuristic it might seem to give the solver supplementary knowledge, thus significantly increasing its scores. This effect should then be visible at least at a statistical level on our long, narrative texts: if the heuristic, as is, has any value for reference resolution, it should improve the scores.

We defined two boolean parameters to model this rule, allowing for four combinations. The parameters are named 'Oblige_MRcreation_for_indefiniteNP' and 'Oblige_association_for_ definiteNP'. Results are given Tables 4 and 5 for the two texts.

| Create an MR for an indefinite NP | Associate a definite NP to an MR | MUC f-measure | Core-MR | | | ExCore-MR |
|---|---|---|---|---|---|---|
| | | | recall | precision | f-measure | f-measure |
| possibly | possibly | **71.90** | **47.37** | **48.97** | **48.16** | **73.77** |
| always | possibly | – 1.01 | – 4.51 | + 1.03 | – 2.01 | + 1.03 |
| possibly | always | – 20.28 | – 7.52 | – 21.94 | – 15.95 | – 21.27 |
| always | always | – 21.66 | – 9.40 | – 23.34 | – 17.56 | – 20.44 |

Table 4. Results for the VA text — Different creation / association criteria according to the definiteness of the NP of the current RE. These criteria fail to bring here the improvement suggested by linguistic analysis

| Create an MR for an indefinite NP | Associate a definite NP to an MR | MUC f-measure | Core-MR | | | ExCore-MR |
|---|---|---|---|---|---|---|
| | | | recall | precision | f-measure | f-measure |
| possibly | possibly | **79.98** | **46.35** | **41.25** | **43.65** | **40.30** |
| always | possibly | – 7.25 | – 1.25 | + 4.45 | + 1.75 | + 3.20 |
| possibly | always | – 5.75 | – 1.81 | – 2.35 | – 2.12 | – 0.38 |
| always | always | – 11.88 | – 2.37 | – 1.83 | – 2.08 | – 0.17 |

Table 5. Results for the LPG text — Same legend as Table 4

Experimental results tend thus to contradict the theoretical considerations. None of the two halves of the heuristic, neither their combination, have brought any score improvement. On the contrary, as far as the three scores agree, there has been a deterioration in the solver's performance; the only positive results come from the 'Oblige_MRcreation_for_indefiniteNP' rule, which increases the number of MR creations, thus counteracting an over-grouping tendency, and slightly increasing precision. When both parts of the heuristic are turned on, the scores are cumulatively decreased.

This somehow counterintuitive result may be due to several reasons. First, our texts are one piece stories, so new discourse referents aren't introduced very often, except for the very beginning of the stories. The definite/indefinite rule isn't applied very often in the cases it was designed for. On the other side, the LFG parser isn't enough robust, so it may have missed or mistaken several definite/indefinite features. In any case, our point here was merely to show how such heuristics can be evaluated, being aware that more robust or higher-knowledge systems will certainly bring more reliable linguistic/statistical conclusions.

### 3.4 Measuring the influence of activation parameters

We already mentioned that the solver had also numerical parameters (§2.3). These are: MR activation on creation, MR re-activation according to the nature of the attached RE (common noun, proper name, pronoun), de-activation according to distance in the text (number of words, of sentences, of paragraphs) and size of the buffer containing the active MRs (sort of short term memory, generally between ten and thirty MRs). However, the assessment method presented above doesn't apply directly to them. Indeed, there is no such notion of discarding a parameter, as they are always supposed to have a numerical value, but this value can of course be adapted to obtain a better score. Unlike the rule assessment, where all possible combinations (or most of them) can be tested, the problem in parameter tuning is to find an optimum in a 9-dimension space. We implemented an optimizer using a simple gradient descent method, which picks a parameter at random, changes it to obtain a better score and picks another one. Improvement by this method is quite long, and doesn't exceed 1-2% (described in Popescu-Belis 1998).

### 4. Conclusion

A coreference solver has been designed and tested on two complex narrative and long texts. On such texts, automatic evaluation methods prove essential. To do this evaluation we used the recall and precision calculation such as proposed in MUC, together with two new and less indulgent methods.

The first part of the evaluation process concerned each of the three selectional constraints of the reference solver. And, as we could foresee it, it showed that the semantic constraint is the most essential and its design is worth improving. The second part concerned some criteria suggested by the linguistic analysis, and showed that these criteria failed to bring a clear improvement in the resolution.

The evaluation paradigm we have presented here seems thus relevant and satisfying as it provides an easy way to test the new heuristics and constraints we may wish to bring to our system.